\documentclass[a4paper,fleqn,numbers]{cas-sc}

\usepackage[authoryear,longnamesfirst]{natbib}

\usepackage{graphicx}%
\usepackage{multirow}%
\usepackage{amsmath,amssymb,amsfonts}%
\usepackage{amsthm}%
\usepackage{mathrsfs}%
\usepackage[title]{appendix}%
\usepackage{xcolor}%
\usepackage{textcomp}%
\usepackage{manyfoot}%
\usepackage{booktabs}%
\usepackage{algorithm}%
\usepackage{algorithmicx}%
\usepackage{algpseudocode}%
\usepackage{listings}%

\usepackage{wrapfig}
\usepackage[font=small,skip=8pt]{caption}

\newcommand{\ScienceConcepts}{\text{F1}}
\newcommand{\Rules}{\text{F2}}
\newcommand{\Debug}{\text{F3}}
\newcommand{\Engineering}{\text{F5}}

\begin{document}
\let\WriteBookmarks\relax
\def\floatpagepagefraction{1}
\def\textpagefraction{.001}

\shorttitle{Chain-of-Thought Prompting + Active Learning (CoTAL)}    

\shortauthors{Cohn et al.}  

\title [mode = title]{CoTAL: Human-in-the-Loop Prompt Engineering for Generalizable Formative Assessment Scoring and Feedback}  




\author[1]{Clayton Cohn}[orcid=0000-0003-0856-9587]
\cormark[1]
\ead{clayton.a.cohn@vanderbilt.edu}

\author[1]{Ashwin T S}[orcid=0000-0002-1690-1626]
\author[1]{Naveeduddin Mohammed}[orcid=0000-0002-3706-2884]
\author[1]{Gautam Biswas}[orcid=0000-0002-2752-3878]

\ead[url]{https://www.vanderbilt.edu/oele/}


\affiliation[1]{organization={Vanderbilt University},
            city={Nashville},
%          citysep={}, % Uncomment if no comma needed between city and postcode
            state={TN},
            country={USA}
}

% \author[2]{}%[]

% Footnote of the second author

% Email id of the second author
% \ead{}



% \affiliation[2]{organization={},
% %          citysep={}, % Uncomment if no comma needed between city and postcode

\cortext[1]{Corresponding author}



\begin{abstract}
Large language models (LLMs) have created new opportunities to assist teachers and support student learning. While researchers have explored various prompt engineering approaches in educational contexts, the degree to which these approaches generalize across domains---such as science, computing, and engineering---remains underexplored. In this paper, we introduce \textit{Chain-of-Thought Prompting + Active Learning} (CoTAL), an LLM-based approach to formative assessment scoring that  (1) leverages Evidence-Centered Design (ECD) to align assessments and rubrics with curriculum goals, (2) applies \textit{human-in-the-loop} prompt engineering to automate response scoring, and (3) incorporates \textit{chain-of-thought} (CoT) prompting and teacher and student feedback to iteratively refine questions, rubrics, and LLM prompts. Our findings demonstrate that CoTAL improves GPT-4's scoring performance across domains, achieving gains of up to 38.9\% over a non-prompt-engineered baseline (i.e., without labeled examples, chain-of-thought prompting, or iterative refinement). Teachers and students judge CoTAL to be effective at scoring and explaining responses, and their feedback produces valuable insights that enhance grading accuracy and explanation quality.
\end{abstract}


\begin{highlights}
\item CoTAL unifies Evidence-Centered Design, human-in-the-loop prompt engineering, chain-of-thought prompting, and active learning in a generalizable framework for formative assessment scoring and feedback.
\item Across science, computing, and engineering assessments, CoTAL improved GPT-4 scoring over a non-prompt-engineered baseline---with gains up to 38.9\% and achieving 94.7\% agreement with human scoring on test responses.
\item CoTAL produced interpretable score explanations by citing evidence from student answers and linking it explicitly to rubric criteria, supporting transparent and rubric-aligned automated assessment.
\item Teacher feedback suggests LLM-based formative assessment can reduce grading burden, surface student misconceptions, and support more targeted instructional intervention without replacing teacher judgment.
\item Student feedback indicates that explanatory AI scoring can build trust and help learners identify mistakes and next steps, highlighting the importance of actionable, learner-centered feedback design.
\end{highlights}

% Keywords

\maketitle

% Main text
\section{Introduction} \label{sec:introduction}

% STEM+C
In K-12 STEM+C (Science, Technology, Engineering, Mathematics, and Computing) classrooms, educators foster engagement by linking scientific principles to real-world phenomena, enabling students to develop exploration, inquiry, and problem-solving skills. This provides students with opportunities to develop a foundational understanding of essential STEM+C concepts and practices \citep{hutchins2023bjet}. Unlike single-discipline curricula, STEM+C learning requires students to integrate cross-domain concepts \textit{synergistically}, linking ideas from one domain (e.g., science) to another (e.g., computing). While this approach enhances learning outcomes and promotes a deeper understanding of scientific processes, it also introduces complexities that can hinder learning \citep{basu2016identifying,hutchins2020c2stem}, necessitating additional guidance and support that formative assessments can provide.

% Assessments
\textit{Formative assessments} employing open-ended questions help students identify key constructs, apply concepts to reasoning and problem-solving tasks, and help develop critical thinking skills \citep{grover2016assessing}. They simultaneously provide teachers with granular insights into students' STEM+C knowledge and problem-solving abilities \citep{CIZEK20231}, allowing them to monitor progress and adapt instruction to student needs. Unlike summative assessments and standardized tests, which are primarily evaluative and often insensitive to the complexities of students' learning challenges \citep{mislevy2003brief}, formative assessments promote self-reflection in students when they face difficulties and facilitate targeted feedback that helps them refine their understanding and improve performance.

% ECD
Evidence-based approaches for generating formative assessments and evaluation rubrics help make explicit their alignment with curricular goals and designated standards \citep{Wise_Shaffer_2015}. Evidence-centered design (ECD) enables a more nuanced and flexible approach by adopting cognitive science and instructional design approaches that embed evidentiary reasoning into each stage of assessment and rubric development \citep{mislevy2003brief}.

% Automated FA grading challenges
However, challenges persist in classroom environments for generating, grading, and providing timely feedback for formative assessments that align with educational standards. Designing and grading free-response formative assessments can place excessive demands on classroom teachers, who may have limited expertise in integrating STEM+C subjects, potentially affecting their ability to create assessments and rubrics that accurately evaluate students' interdisciplinary knowledge \citep{hutchins2023bjet}. Students may also lack mature writing skills, necessitating significant time and effort from human graders to infer the true implications of students' answers \citep{sari2021effect}. Consequently, research is needed to develop automated scoring systems that efficiently deliver needed feedback to facilitate STEM+C learning based on teacher preferences \citep{burrows2015eras}.

% Prompt engineering
Prompt engineering with large language models (LLMs) offers a promising solution, allowing users to adapt language models to downstream tasks by incorporating approaches like \textit{in-context learning} (ICL) \citep{brown2020language}, \textit{chain-of-thought} (CoT) prompting \citep{wei2022chain}, and \textit{active learning} \citep{cohn1994improving} that obviate the need for traditional training via parameter updates, thus conserving data and computational resources often unavailable in educational settings \citep{cochran2023bimproving}. ICL enables LLMs to ``learn''\footnote{We use the term ``learn'' to indicate that the LLM uses examples to guide its predictions, not to suggest that ICL involves learning via parameter updates, which it does not.} from \textit{few-shot} examples in the prompt during inference. CoT extends ICL by augmenting labeled few-shot examples with step-by-step reasoning chains to provide more explicit guidance to the LLM for scoring student answers \citep{golchin2024grading,impey2025using}. Active learning is a \textit{human-in-the-loop} approach to improving model training, where the human serves as an \textit{oracle}, selecting additional instances to label for the next training iteration to enhance performance and robustness \citep{cohn1994improving}. 

% Our work
In this paper, we leverage human-in-the-loop prompt engineering to realize a \textit{stakeholder-AI partnership}\footnote{For this paper, ``stakeholders'' refers to teachers, students, and researchers.}, enriching assessment and feedback within an NGSS-aligned (Next Generation Science Standards; \cite{ngss2013}) Earth science curricular unit \citep{mcelhaney2020isls,zhang2020aied}. Building on our previous work on automated scoring for short-answer science assessments \citep{cohn2024chain}, we propose a \textit{generalizable} approach: \textit{Chain-of-Thought Prompting + Active Learning} (CoTAL) that employs LLMs to automate scoring and feedback for formative assessments across multiple domains. We define generalizability as the ability to grade formative short-answer responses varying by question type (e.g., concept definitions, process descriptions, comparisons, explanations), rubric structure (e.g., multi-label vs. multi-class), and content domain (science, computing, engineering).

% RQs
To evaluate CoTAL, we address the following research questions:

\begin{itemize}
    \item \textbf{RQ1}. Can CoTAL improve an LLM's ability to score and explain responses to formative assessment questions across multiple connected domains?
    \item \textbf{RQ2}. What do teacher and student input reveal about the effectiveness, actionability, and impact of CoTAL's formative feedback?
\end{itemize}

% Answering RQs
We answer RQ1 using a mixed-methods approach. First, we conduct a quantitative evaluation of CoTAL's scoring performance using GPT-4 on formative assessment questions in science, computing, and engineering, comparing it to a non-prompt-engineered baseline. We use Cohen's Quadratic Weighted Kappa (QWK; \cite{cohen1968weighted}) to measure agreement and assess CoTAL's impact on scoring accuracy. Second, we qualitatively analyze GPT-4's reasoning chains by performing a constant comparative analysis \citep{charmaz2006} to identify the strengths and weaknesses of the LLM's scoring justifications when using CoTAL. We answer RQ2 qualitatively by conducting interviews with teachers and surveying students to assess the classroom effectiveness of LLM-generated formative feedback. We memo key findings from the teacher interviews \citep{hatch2002} and apply constant comparative analysis to student survey responses, using stakeholder input to guide iterative methodological and curricular refinements.

\section{Background} \label{sec:background}

\subsection{Evidence-Centered Design}

ECD structures assessments around evidentiary reasoning. Domain, Evidence, and Task models specify the knowledge, skills, and abilities (KSAs) aligned with standards (e.g., NGSS \citeyearpar{ngss2013}) and expertise; identify observable indicators of mastery; and design activities that elicit the required evidence \citep{mislevy2003brief}. This alignment supports formative assessments and rubrics that faithfully track student understanding against curricular goals.

LLMs enable evidence elicitation during automated formative assessment grading via CoT prompting. Beyond assigning scores, LLMs can quote student responses, tie them to assessment questions and rubric items, and pinpoint areas of proficiency and difficulty. This alignment grounds LLM reasoning, minimizes hallucinations, and supports accurate scoring and mastery evidence generation.

While substantial work has explored ECD-driven formative assessment \citep{kubsch2022toward,muftuouglu2024framework,mislevy2003brief}, few studies have applied ECD principles to automating formative assessment scoring and feedback. \cite{hao2025ai} takes a critical first step by examining the challenges and opportunities of integrating ECD into LLM systems, advocating for ``continuous research, critical evaluation, and open collaboration to ensure responsible, effective integration'' of LLMs into assessment practices. However, this work stops short of proposing a concrete implementation. \cite{zapata2024designing} investigate the integration of LLMs and ECD for conversation-based assessment scoring, explicitly calling for human-in-the-loop approaches to mitigate fairness and bias issues in LLMs. Their study focused primarily on the prompt engineering process and did not evaluate ECD-driven LLM feedback with classroom learners.

\subsection{Automated Formative Assessment Scoring and Feedback}

Previously, automated assessment scoring methods have incorporated data augmentation \citep{cochran2022improving,cochran2023improving,zhang2025data}, next sentence prediction \citep{wu2023matching}, domain adaptation and supervised fine-tuning \citep{cochran2023using}, prototypical neural networks \citep{zeng2023generalizable}, cross-prompt fine-tuning \citep{funayama2023reducing}, and reinforcement learning \citep{liu2022giving} to improve grading accuracy. However, these methods have largely focused on more structured grading tasks in mathematics and computer science \citep{nakamoto2023enhancing,paiva2022automated}, where open-ended responses are less prevalent than in science domains \citep{cohn2024chain}. These approaches have achieved varying degrees of success but often fail to provide comprehensive insights into their scoring decisions.

% Grading
More recently, researchers have explored the use of LLMs for automated scoring in science \citep{villagran2024implementing}. \cite{golchin2024grading} adapted zero-shot chain-of-thought prompting to grade writing assignments in Coursera courses on Astronomy and Astrobiology, benchmarking GPT-4 against instructor and peer evaluations. \cite{impey2025using} assessed GPT-4's rubric-based grading reliability on similar MOOC assignments, finding its scores closely aligned with both instructors and peers. While \cite{impey2025using} caution that rubric interpretation can vary between humans and GPT-4 and note that findings from low-stakes MOOC environments may not generalize to high-stakes classroom settings, Golchin et al. (2024) report degraded performance on tasks requiring creative or speculative reasoning---underscoring the need for human-in-the-loop evaluations conducted in situ. \cite{pinto2023large} used ChatGPT to grade open-ended questions in technical training, experimenting with different levels of determinism (i.e., temperature settings). However, their study was conducted strictly in a zero-shot setting and focused on adult technical training rather than K-12 STEM learners.

% Feedback
Researchers have also begun to leverage the rich feedback capabilities offered by LLMs. \cite{matelsky2023large} present a domain-agnostic, open-source widget that uses LLMs to automate personalized feedback on students' open-ended responses. However, without human validation, the tool's effectiveness hinges on the quality and completeness of the criteria provided by the user (i.e., the instructor), and it has not yet been empirically validated in real-world classrooms. \cite{meyer2024using} used LLMs to bring evidence-based feedback into the classroom, investigating whether AI-generated feedback improves secondary students' essay revisions, motivation, and emotions compared to no feedback. However, their study focuses on English-as-a-foreign-language students rather than STEM learners.

\subsection{Human-in-the-Loop Prompt Engineering}

Unlike algorithmic prompt engineering, human-in-the-loop approaches \citep{cohn2024chain,cohn2024human} integrate an important human perspective into the design and optimization pipeline that enables users to influence LLM outputs rather than relying solely on algorithmic decisions \citep{ranade2024using,cohn2024human}. This creates a human-AI collaboration \citep{cohn2024multimodal,jarvela2025hybrid} that enhances LLM alignment with user preferences and ensures the contextual relevance of the generated output by incorporating nuanced domain knowledge that algorithmic methods may overlook. Furthermore, this approach helps mitigate LLM errors and hallucinations by leveraging human expertise to validate and refine generated responses, fostering more accurate, ethical, and trustworthy LLM interactions. 

Beyond \cite{cohn2024chain,cohn2024human}, \cite{lee2024applying} showed that CoT prompting with contextual item stems and rubrics improves LLM scoring of middle school science assessments by emphasizing domain-specific reasoning. Their human-in-the-loop procedure, WRVRT (Writing, Reviewing, Validating, Revising, Testing), combines CoT prompting with iterative refinement to enhance accuracy and explainability. However, their work is limited to science, leaving cross-subject generalizability untested; moreover, it centers on prompt engineering without incorporating ECD principles or integrating teacher and student feedback to refine assessments, rubrics, or prompts.

\subsection{Active Learning}
The term \textit{active learning} is formally defined in both educational and AI contexts. In education, active learning is defined by \cite{bonwell1991active} as ``\textit{instructional activities involving students in doing things and thinking about what they are doing,}'' emphasizing a shift from passive information reception to skill development through meaningful engagement (e.g., group work and projects) that fosters discussion and reflection. 

In AI, \cite{cohn1994improving} define active learning as ``\textit{any form of learning [i.e., training] in which the learning program has some control over the inputs on which it trains.}'' Rather than passively consuming labeled examples, active learning algorithms selectively query a human ``oracle'' for labels on the most informative instances to optimize the training process. This approach ensures models are trained on impactful data, learning from fewer examples and inducing more precise decision boundaries.

In our work, we adopt the AI definition of active learning, establishing it as a human-in-the-loop prompt engineering approach to iteratively refine prompts and augment them with few-shot examples to improve in-context learning by: (1) testing prompts on a validation set; (2) identifying trends in scoring deficiencies (i.e., factors causing LLM misalignment with human judgment); and (3) incorporating targeted few-shot examples using chain-of-thought (CoT) prompting to address those deficiencies and correct misalignment, following the approach introduced by \cite{cohn2024chain}. 

Other researchers have proposed a related strategy, i.e., \textit{active prompting} \citep{qian2024ape,diao2024active}, which leverages uncertainty metrics (e.g., entropy or margin) to identify examples with the least confident predictions. The top-$k$ most ambiguous cases are sent to human annotators for labeling or corrective feedback, which are then incorporated directly into the prompt as demonstrative examples. However, unlike active learning, active prompting emphasizes aggregate uncertainty rather than targeting specific misalignment patterns responsible for LLM disagreement. In contrast, our work addresses specific LLM misalignments in cases where discernible trends emerge during validation set evaluation, as elaborated in Section \ref{subsec:cot_al}.

\subsection{Research Gaps}

While researchers have employed a variety of prompt engineering techniques to improve LLM performance on formative assessment scoring \citep{li2024autograding}, these approaches typically focus on a single domain and dataset, leaving their generalizability largely unevaluated. Moreover, they often neglect stakeholder input in the development and refinement of their methods. A notable gap remains in automated formative assessment scoring approaches that (1) integrate ECD principles with prompt engineering, (2) provide transparent explanations of scoring decisions to teachers and students, (3) demonstrate effectiveness across multiple domains, and (4) incorporate stakeholder feedback for iterative refinement.

In this paper, we address these gaps by proposing a generalizable, stakeholder-informed framework that combines ECD with human-in-the-loop prompt engineering (i.e., active learning) to enable transparent, accurate, and adaptable formative assessment scoring. Our work builds on prior research by explicitly grounding our methodology in assessment design theory, engaging both teachers and students in the refinement process, and evaluating performance across science, engineering, and computing domains in real-world classroom settings.

\section{SPICE Curriculum} \label{sec:spice_curriculum}

SPICE (Science Projects Integrating Computing and Engineering) is a three-week middle school curriculum unit in which students redesign a schoolyard using surface materials that minimize post-rainstorm runoff while satisfying design constraints (e.g., cost, accessibility). The problem-based curriculum comprises five units: (1) physical experiments; (2) conceptual modeling; (3) paper-based computational thinking tasks; (4) computational modeling of water runoff; and (5) engineering design using the computational model to evaluate solutions. Students build computational models in SPICE's computer-based environment and apply them to redesign the schoolyard under engineering constraints \citep{hutchins2020domain}. The curriculum targets NGSS performance expectations for upper elementary and middle school Earth science and engineering design, emphasizing surface water movement and human impact.

Formative assessments are interspersed throughout the curriculum to help students self-evaluate their learning progress. These assessments help teachers monitor and support students' science, computing, and engineering learning as they progress through the curriculum. The associated grading rubrics (discussed shortly) reflect cross-domain connectivity, enabling teachers to monitor students' progress and modify instruction when students have difficulties.

\begin{figure}
    \centering
    \includegraphics[width=.9\columnwidth]{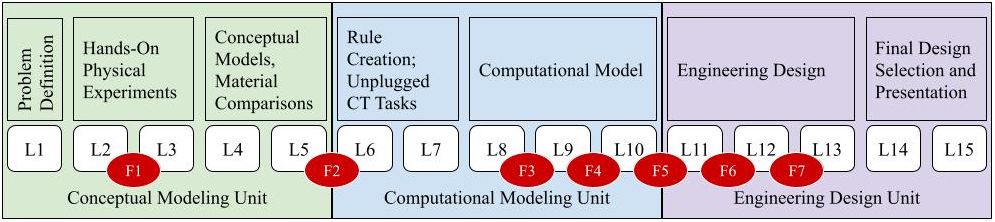}
    \caption{SPICE curricular sequence (L items in white are lessons; F items in red are formative assessments. ``CT'' refers to computational thinking, which we use synonymously with ``computing'' for the purposes of this paper.}
    \label{fig:spice-assessments}
\end{figure}

SPICE leverages ECD to systematically create assessments and tasks to evaluate student learning in science, computing, and engineering. For our analysis, we picked three of the seven formative assessments: (1) \Rules  (conceptual modeling in science); (2) \Debug  (computational modeling of science phenomena); and (3) \Engineering  (testing engineering designs), developed to monitor and support students' learning across the three domains in the curriculum. The locations of the assessments in the integrated curriculum are shown at the identified markers in Figure \ref{fig:spice-assessments}. 

Previous work focused on the automated grading of \textit{Science Concepts and Reasoning Task} assessments from the science unit (\ScienceConcepts). This paper generalizes the previous approach to include three additional formative assessments that cover: 
\begin{enumerate}
    \item a \textit{Rules Task} (\Rules) that requires students to structure their understanding of conservation of matter by expressing the relation between the amounts of rainfall, water absorbed by the surface material, and runoff as three separate ``rules'';
    \item a \textit{Debugging Task} (\Debug) requiring students to check the conditional form of the rules and value expressions from the \textit{Rules Task} by analyzing block-based code generated by a fictional classmate to generate a computational science model of the runoff phenomena and identifying errors in the computational model; and
    \item an \textit{Engineering Task} (\Engineering) where students integrate their knowledge of science and computing concepts with the engineering principle of \textit{fair tests} to ensure a fair comparison between two designs (i.e., the conditions under which the designs are evaluated remain consistent).
\end{enumerate} 

These assessments (and their rubrics) were explicitly designed for students to realize the connections between the science, computing, and engineering domains and to form a cumulative understanding of cross-domain conceptual knowledge as they progress through the curriculum \citep{hutchins2021isls,zhang2020aied}.

The \textit{Rules Task} nudges students to translate their learned intuitions about the conservation of matter principle into a quantitative relation between variables (total rainfall, total absorption, absorption limit, and total runoff). Students express this relation by considering three scenarios, i.e., when rainfall is greater than, less than, and equal to the surface absorption limit. Students are then asked to express these three scenarios as conditional logic expressions defining absorption and runoff values. For example, if \textit{rainfall}$<$\textit{absorption limit} of the ground material, then \textit{absorption = rainfall} and \textit{runoff = 0}. Later, students must recall these conditional logic expressions to construct their computational models. We identify the \textit{Rules Task} rubric as categorical, as there is a specific structure to the response, and students receive 1 point for including each required component per rule. The rubric for the \textit{Rules Task} appears in Table \ref{tab:rules_rubric}.

\begin{table}
    \centering
    \begin{tabular}{| p{0.12\linewidth} | p{0.60\linewidth} | p{0.11\linewidth} |}
        \hline
        \textbf{Subscore} & \textbf{Description} & \textbf{Domain} \\
        \hline
        R1 & Completed if statement ``if rainfall is $<$ absorption limit.'' & SCI, COM \\
        R2 & Set absorption to rainfall in this rule. & SCI \\
        R3 & Set runoff to 0 in this rule. & SCI \\
        \hline
        R4 & Completed if statement ``if rainfall $=$ to absorption limit.'' & SCI, COM \\
        R5 & Set absorption to rainfall OR absorption limit in this rule. & SCI \\
        R6 & Set runoff to 0 in this rule. & SCI \\
        \hline
        R7 & Completed if statement ``if rainfall $>$ than absorption limit.'' & SCI, COM \\
        R8 & Set absorption to absorption limit in this rule. & SCI \\
        R9 & Set runoff to ``rainfall - absorption limit'' OR ``rainfall - absorption'' in this rule. & SCI \\
        \hline
    \end{tabular}
    \caption{Categorical rubric used for the \textit{Rules Task}. Each ``R'' corresponds to a different subscore for the \textit{Rules Task} and is explained in the table. The \textit{Rules Task} targets the science (SCI) and computing (COM) domains (with an emphasis on science).}
    \label{tab:rules_rubric}
\end{table}

Table \ref{tab:rules_rubric} enumerates the nine possible points (subscores) for the \textit{Rules Task} (R1-R9). Students receive 1 point per correct conditional statement they identify (R1, R4, R7). Within each conditional statement, students receive 1 point for correctly setting the absorption value (R2, R5, R8) and 1 point for correctly setting the runoff value (R3, R6, R9). For example, the statement ``\textit{if rainfall is equal to the absorption limit, then set absorption to rainfall, and set runoff to zero}'' would earn 3 points (R4, R5, and R6).

\begin{figure}
    \centering
    \includegraphics[width=.97\columnwidth]{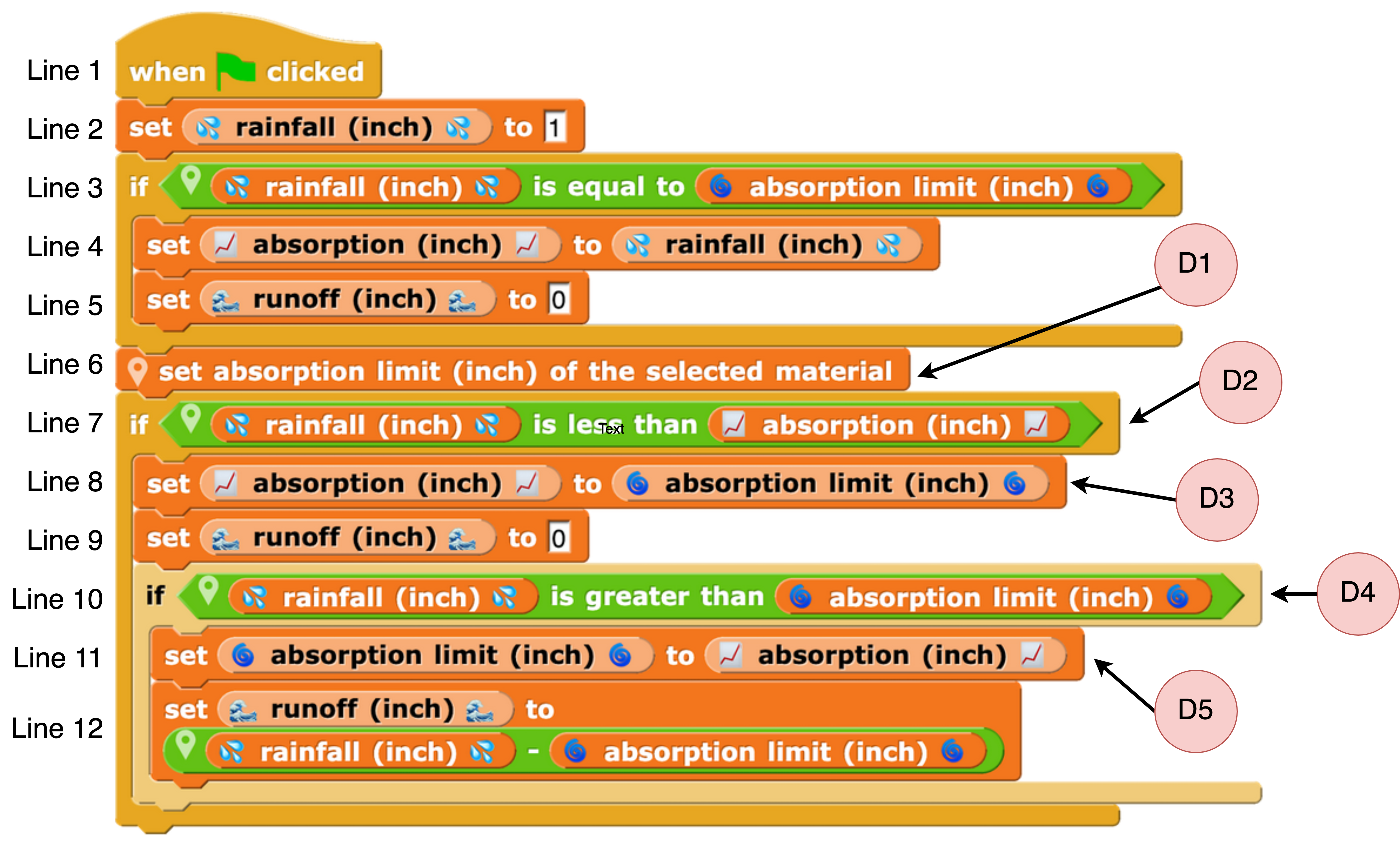}
    \caption{Erroneous code presented to students during the \textit{Debugging Task}. Red ``D'' circles correspond to the individual model errors presented in Table \ref{tab:debug_rubric}.}
    \label{fig:debug_code}
\end{figure}

For \Debug, the \textit{Debugging Task}, students are asked to identify and describe the five errors present in a fictional student's code (illustrated in Figure \ref{fig:debug_code}). To accomplish this, students must understand the conditional logic expressions they wrote for the \textit{Rules Task} and accurately translate them into block-structured programming code using ``if statements'' and expressions in the SPICE environment to model the amount of absorbed water and runoff. Students learn how to use the coding blocks to build their computational water runoff models before they work on the \textit{Debugging Task}. The rubric for this task is shown in Table \ref{tab:debug_rubric}.

\begin{table}
    \centering
    \begin{tabular}{| p{0.12\linewidth} | p{0.60\linewidth} | p{0.11\linewidth} |}
        \hline
        \textbf{Subscore} & \textbf{Description} & \textbf{Domain} \\
        \hline
        D1 & ``Set absorption limit'' should be before the first conditional statement. & COM \\
        \hline
        D2 & In the ``less than'' condition, rainfall should be compared to the absorption limit. & SCI, COM \\
        \hline
        D3 & In the ``less than'' condition, absorption should be set to rainfall. & SCI \\
        \hline
        D4 & The ``greater than'' condition should not be embedded in the less than condition, but connected to it. & COM \\
        \hline
        D5 & In the ``greater than'' condition, absorption should be set to absorption limit, not the other way around. & SCI, COM \\
        \hline
    \end{tabular}
    \caption{Categorical rubric used for the \textit{Debugging Task}. Each ``D'' corresponds to a different subscore (i.e., code error; see Figure \ref{fig:debug_code}) for the \textit{Debugging Task} and is explained in the figure. Like the \textit{Rules Task}, the \textit{Debugging Task} targets the science (SCI) and computing (COM) domains (but with an emphasis on computing).}
    \label{tab:debug_rubric}
\end{table}

Table \ref{tab:debug_rubric} shows that students can earn up to 5 points (subscores; one for each error they identify) for the \textit{Debugging Task}. ``D'' values refer to the individual errors the students must identify in the model. D1 refers to the ``set absorption limit (inch) of the selected material'' block being erroneously placed on line 6 (it should come before the first conditional statement on line 3). D2 refers to rainfall being incorrectly compared to absorption in the ``less than'' condition on line 7 (it should be compared to the absorption limit). D3 refers to absorption incorrectly being set to the absorption limit inside the ``less than'' condition on line 8 (absorption should be set to rainfall). D4 refers to the ``greater than'' condition being improperly set on line 10 (it should not be nested inside the ``less than'' condition). D5 refers to the absorption and absorption limit being swapped inside the ``greater than'' condition on line 11 (absorption should be set to absorption limit, not the other way around).

The \textit{Engineering Task} formative assessment (\Engineering) requires students to integrate their science and computing domain knowledge with their engineering knowledge of design constraints and fair tests. Students compare two design solutions generated by a fictional student and are provided information about each design test's input (e.g., rainfall) and output (e.g., cost and runoff). Students are then asked to explain whether the provided information allows them to conclude that one design is better than the other.

This task utilizes a new rubric structure where students are awarded a single numerical score from 0 to 4 points. Students are assessed on their ability to determine if the reported tests allow a valid comparison between the two design solutions. Since the fictional student uses different rainfall values (i.e., inputs) to compare the runoff between the designs, the comparison is not ``fair'' because the outcome variable is not generated using the same input for both tests. The students' explanations are evaluated at different levels using the rubric in Table \ref{tab:engineering_rubric}.  

\begin{table}
    \centering
    \begin{tabular}{| p{0.07\linewidth} | p{0.62\linewidth} | p{0.15\linewidth} |}
        \hline
        \textbf{Score} & \textbf{Description} & \textbf{Domain} \\
        \hline
        4 & Student explains that (1) the designs cannot be compared because different rainfall values were used to test each one, and (2) the runoff comparisons will not be ``fair.'' & ENG, SCI, COM \\
        \hline
        3 & Student explains the designs cannot be compared because different rainfall values were used to test each one. & ENG, SCI \\
        \hline
        2 & Student explains the designs cannot be compared because both tests violate design constraints, demonstrating an understanding of constraint satisfaction but not the need for fair tests. & ENG \\
        \hline
        1 & Student identifies that the designs cannot be compared but does not provide reasoning. & ENG \\
        \hline
        0 & Student answers ``yes'' that both designs can be compared fairly. & SCI, COM \\
         \hline
    \end{tabular}
    \caption{\textit{Engineering Task} rubric, targeting the science (SCI), computing (COM), and engineering (ENG) domains (with an emphasis on engineering).}
    \label{tab:engineering_rubric}
\end{table}

Table \ref{tab:engineering_rubric} shows that students are awarded 0 points if they fail to recognize that the two designs cannot be compared fairly (i.e., they answer ``yes'' to the question posed). Students receive one point if they identify that the two tests cannot be compared fairly (i.e., they answer ``no'' to the question posed) but do not provide a meaningful explanation. Students receive two points if they discuss design constraints as a reason that the two tests are not comparable. Three points are awarded if the students discuss the differing rainfall values as the reason why the tests cannot be compared fairly. Four points are awarded if the students mention the differing rainfall values and explain that this results in unfair runoff comparisons. 

\section{Methods}

\subsection{Chain-of-Thought Prompting + Active Learning (CoTAL)}\label{subsec:cot_al}

\begin{figure}
    \centering
    \includegraphics[width=\textwidth]{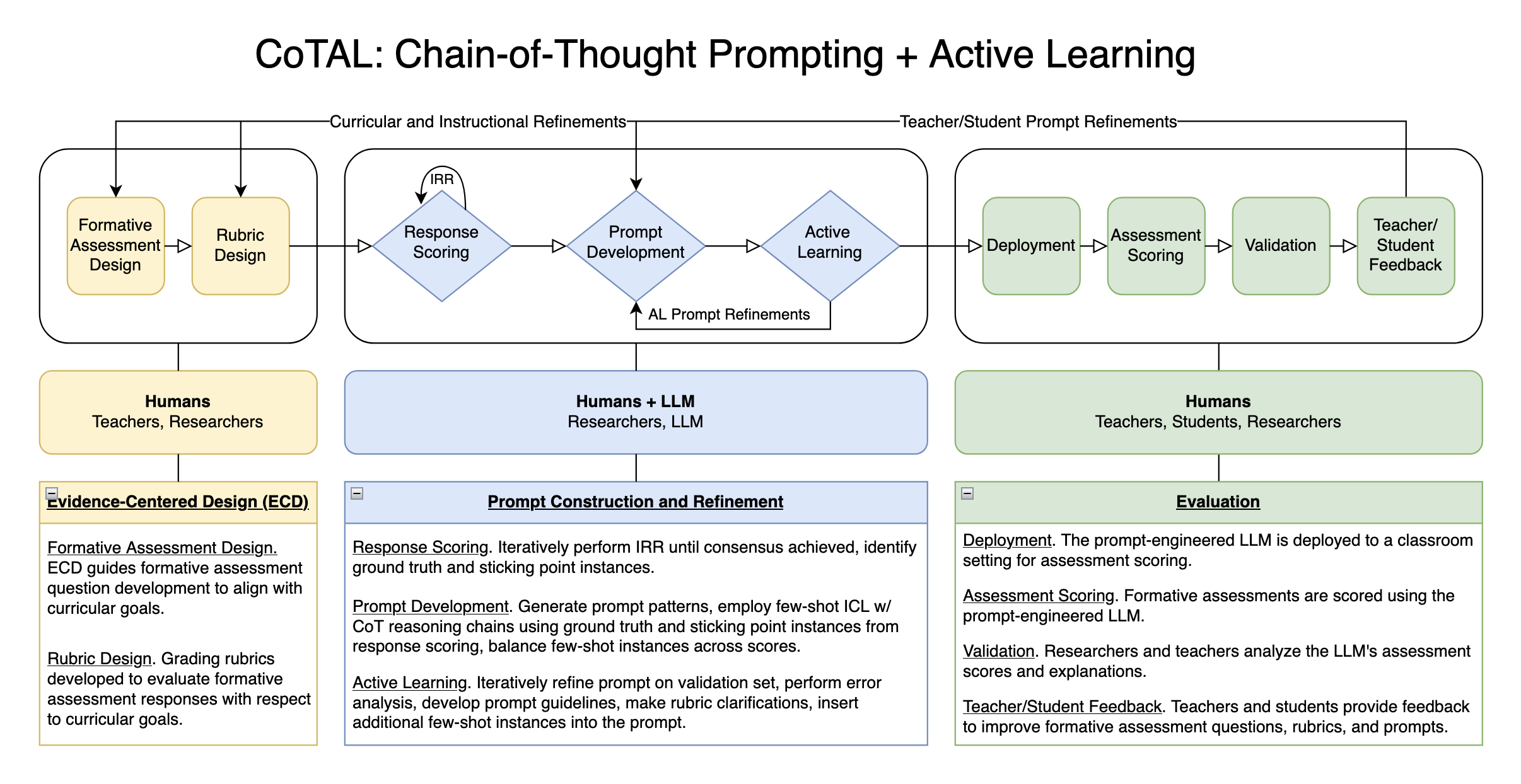}
    \caption{Chain-of-Thought Prompting + Active Learning (CoTAL).}
    \label{fig:method}
\end{figure}

We have developed CoTAL (illustrated in Figure \ref{fig:method}), as a generalizable method for improving automated formative assessment scoring and aligning LLM scores and explanations with the needs of teachers and students. Our approach fosters a collaborative partnership among researchers, teachers, students, and AI, integrating ECD, human-in-the-loop prompt engineering, and stakeholder feedback to (1) score and explain students' short-answer responses to formative assessment questions in the SPICE curriculum; and (2) iteratively refine our formative assessments, rubrics, and grading prompts based on student and teacher feedback. 

CoTAL comprises three phases, as illustrated in Figure \ref{fig:method}. \textbf{Phase I} (yellow) is human-driven: teachers and researchers apply ECD to develop curriculum-aligned formative assessments and rubrics, guided by the evidence needed to identify student mastery relative to curricular goals, and enabling effective evaluation of student responses (Section  \ref{sec:spice_curriculum}). The LLM is not involved in Phase I. \textbf{Phase II} (blue) co-develops and optimizes an initial prompt with the LLM by: 
\begin{enumerate} 
    \item \textit{Response Scoring}, which samples student responses, conducts IRR (e.g., Cohen's Kappa; \cite{cohen1960coefficient}), and resolves disagreements to establish consensus; 
    \item \textit{Prompt Development}, which employs few-shot \textit{CoT} reasoning aligned to human consensus and structures the prompt with targeted \textit{prompt patterns}; and
    \item \textit{Active Learning}, which iteratively refines the prompt on a validation set to align generations with human scoring preferences.
\end{enumerate} 
These map to the three blue diamonds in Figure \ref{fig:method}. Phase III deploys the refined prompt in classrooms to score responses and generate explanations for assigned scores. The researchers then use teacher and student critiques to inform targeted refinements to questions, rubrics, and prompts without degrading LLM scoring performance. This procedure extends prior work on the \textit{Science Concepts and Reasoning Task} \citep{cohn2024chain} to three integrated assessments in science, computing, and engineering (Section \ref{sec:spice_curriculum}).

During \textit{Response Scoring} (Phase II, Step 1), experts independently score a random subset of student responses with Phase I rubrics, and establish their initial degree of agreement based on a Cohen's $\kappa$. When discrepancies arise between the scores assigned by the two reviewers, the underlying reasons for these differences are analyzed and resolved to reach a final consensus. Difficult-to-resolve discrepancies, termed ``\textit{sticking points},'' receive special attention during Prompt Development (Phase II, Step 2) to refine the prompt structure and guide the LLM toward alignment with the human scorers. Sticking points often emerge from ``edge cases'' where the reviewers interpret the same rubric differently. For example, one \textit{Engineering Task} sticking point was whether a general understanding of design constraints (i.e., the student did not explicitly mention any constraint by name) qualified for 2 points (both reviewers ultimately agreed this was acceptable). This process---randomly sampling the data, calculating Cohen's $\kappa$, and documenting reasons for disagreement---is repeated until $\kappa \geq 0.70$ to ensure consensus. This is depicted as the ``IRR'' self-loop in Figure \ref{fig:method}. 

During \textit{Prompt Development} (Phase II, Step 2), the LLM is first given task instructions, followed by the formative assessment question, grading rubric, and additional context to guide the LLM's scoring. This supplementary information helps the LLM connect student responses to the rubric and comprehend the intended concepts, questions, and scoring criteria. Additionally, various \textit{prompt patterns} \citep{white2023prompt} are employed to guide the LLM during inference and help structure its output, including a: (1) \textit{persona pattern} that assigns the LLM a ``persona,'' or role, to play while generating its output; (2) \textit{context manager pattern} that defines the context the LLM should consider while generating its output; (3) \textit{template pattern} that provides the LLM with a structured output template; and (4) \textit{meta language creation pattern} that creates a custom language for the LLM to understand (e.g., a textual shorthand notation to help the LLM interpret graphs) \citep{white2023prompt}. For example, the context manager pattern helped the LLM understand the relationship between the \textit{Rules} and \textit{Debugging Tasks} (i.e., the students often referred to conditional statements by ``rule'' number during the \textit{Debugging Task}), and the meta language creation pattern allowed us to distill the \textit{Debugging Task}'s computational model image into LLM-readable text.

After constructing the prompt, few-shot instances are selected, equipped with CoT reasoning chains, and appended to the prompt to align the LLM with the consensus reached by human scorers. Each formative assessment prompt features two types of few-shot examples: (1) \textit{ground truth} instances, where both scorers agreed during Response Scoring; and (2) \textit{sticking point} instances, where scorers disagreed during Response Scoring, and the reasons for disagreement carried over to other instances, leading to similar disagreements among the scorers. For \textit{sticking point} instances, CoT reasoning chains help clarify potential misunderstandings and guide the LLM toward human consensus. All few-shot instances are accompanied by CoT reasoning chains that loosely (i.e., not verbatim) follow this template:
\begin{quote}
    \small
    \textit{The student says X. The rubric states Y. Based on the rubric, the student earned a score of Z.}
\end{quote}

Unlike traditional CoT prompting, which relies on the LLM to generate intermediate reasoning chains based solely on patterns learned during training, CoTAL grounds the LLM's responses by instructing it to cite relevant portions of students' answers verbatim and link them to the scoring criteria in the rubric. In this way, the LLM's scoring decisions and explanations are inherently guided by human input during inference, ensuring generations remain faithful to the established criteria.

For instance, consider the following assessment question, rubric item, student response, and chain-of-thought from our Engineering Task (FA5) prompt:\footnote{This example is used as a few-shot exemplar in the \textit{Engineering Task} prompt. Prompts and demonstration instances for all three formative assessments are provided in our Supplementary Materials: \url{https://github.com/claytoncohn/CEAI26_Supplementary_Materials}.}

\begin{quote}
    \textbf{Question}: A fictitious student, Morgan, has created two different designs. Morgan wants to test both of her designs to see which one is better [both are presented graphically]. Can Morgan tell which design is better based on these tests? Explain why or why not.\\
    \textbf{Rubric Item}: 4 points: The student indicates the two designs cannot be compared because Morgan used different rainfall values to test each design AND the student explains that the tests are not fair (because runoff cannot be equally compared for the two tests if different amounts of rainfall are used). The student does not have to use the word ``fair'' explicitly to receive credit. He or she can indicate that the tests are not fair by mentioning that the two tests are uneven, inconsistent, impossible to compare, etc. \\
    \textbf{Student Response}: No. Morgan has different inches of rainfall, which means that it is not equal or fair.\\
    \textbf{Chain-of-Thought}: The student indicates he or she understands the two designs cannot be compared by providing an Answer of `No'. Additionally, the student says, `Morgan has different inches of rainfall', which demonstrates the student understands Morgan used different rainfall values to test each design. The student also mentions that `it is not equal or fair', which indicates the student understands that the tests are not fair. The student's response, therefore, meets the Rubric criteria for a maximum score of 4 points. Based on the Rubric, the student earned a score of 4.\\
\end{quote}

In addition to incorporating \textit{ground truth} and \textit{sticking point} examples in the prompt, additional instances are included to ensure that few-shot examples are proportionally represented across subscores to achieve data balance. \cite{min2022rethinking} recommend balancing few-shot instances based on the true distribution of the dataset's labels rather than doing so uniformly. However, in multi-label datasets such as ours, adding individual instances can shift the label distribution for each subscore category, making perfect balance difficult to achieve. At a minimum, few-shot examples should include at least one positive and one negative instance for each subscore (multi-label) or one instance of each score (multi-class).

Once the prompt is constructed, Active Learning (Phase II, Step 3) tests the prompt against a validation set. The LLM's generations are analyzed by first isolating the incorrectly scored instances. For each subscore that the LLM predicts incorrectly, we identify a ``scoring trend'' to determine whether the LLM tends to produce false positives (FPs) or false negatives (FNs) and qualitatively discern the reasons behind these inaccuracies. This process mirrors Response Scoring, where recurring LLM errors on the validation set are identified as \textit{sticking points}. We then select validation set instances exemplifying these \textit{sticking points}, annotating them with CoT reasoning chains to correct the LLM's mistakes and adding them to the existing few-shot examples in the prompt (illustrated by the ``AL Prompt Refinements'' loop in Figure \ref{fig:method}). 

Phase II of CoTAL shares similarities with explainable AI (XAI) approaches, which focus on explaining a model's decisions based on its internal mechanisms. However, in CoTAL's case, the explainability comes from the LLM's \textit{generations}, offering insights into how the model evaluates student responses in relation to rubrics. The emphasis is on alignment with grading criteria rather than interpreting the model's internal logic. This explainability is particularly valued by students, who see it as a critical factor for building trust in AI systems in educational settings (see Section  \ref{sec:rq2_findings}). Unlike our prior work, which primarily highlighted our human-in-the-loop prompt engineering contributions, this paper ``closes the loop'' (see Figure \ref{fig:method}) by (1) integrating LLM prompt engineering with ECD principles, (2) investigating our method's practical utility in classrooms through studies involving both teachers and students, and (3) leveraging student and teacher feedback to inform refinements to our formative assessments, rubrics, and prompts.

In this work, we refine our original Phase II approach based on prior findings \citep{cohn2024chain}. Previously, we conducted active learning by inserting several validation instances back into the prompt. However, we found that the LLM tended to overfit when the number of few-shot instances was large, or the CoT reasoning chains became too granular. In this study, to mitigate the LLM's tendency to overfit, we insert only \textbf{a single instance} into the prompt during Active Learning, specifically targeting the most persistent LLM errors. Additionally, we use Prompt Development (Phase II, Step 2) to provide the LLM with a list of ``guidelines''\footnote{In the actual prompt, we refer to the guidelines as ``rules,'' but we use the term ``guidelines'' in the manuscript so as not to be confused with the ``rules'' in the \textit{Rules Task}.} that the LLM is instructed to adhere to at all times. For example, one \textit{Rules Task} guideline instructed the LLM not to consider the order in which students listed the three rules in its scoring decisions. Human scorers developed these guidelines based on the consensus they reached during the Response Scoring process. Just as we use CoT reasoning to cite evidence from a student's response and connect it to the rubric for scoring, we also apply CoT reasoning to reference these guidelines in order to enhance the responses generated by the LLMs.

In Phase III, the refined prompt from Phase II is implemented in a classroom setting to evaluate students' responses to formative assessments and provide explanations for the assigned scores. Researchers then sample the responses generated by the LLM and present them to teachers for critique. The teachers assess the model's scoring accuracy, soundness of its explanations, and clarity of its responses. During these discussions, the strengths and weaknesses of the language model are identified. Based on the feedback received, the teachers and researchers collaborate to address the model's shortcomings while maintaining its strengths. They agree on specific refinements for the formative assessment questions, scoring rubrics, and prompts.

Similarly, students are shown the LLM's responses to their formative assessment answers and are asked to critique them. While teacher feedback informs methodological improvements related to curricular goals, student feedback highlights user experience and personalized learning. For example, students often emphasize elements such as the LLM's tone in its responses and the overall effectiveness of the content it provides to enhance their understanding of relevant concepts. This process is illustrated in Figure \ref{fig:method} by the ``Curricular and Instructional Refinements'' and ``Teacher/Student Prompt Refinements'' loops. Like Phase I, Phase III depends on human input, involving close collaboration among researchers, teachers, and students to guide curricular and methodological enhancements. 

In this study, we conducted Phase I (formative assessment and rubric development) as a participatory design effort between teachers and researchers. Phase II (prompt engineering) was carried out by members of our research team. Phase III involved continued participatory design with researchers, teachers, and students. This study presents findings on CoTAL's scoring accuracy and its ability to generate scoring explanations (RQ1; Phase II), as well as insights from teachers and students to inform instructional and methodological refinements (RQ2; Phase III). In future work, we will examine the extent to which our Phase III refinements improve the system's ability to meaningfully score student formative assessment responses and provide effective feedback to both students and teachers. The experimental procedure for this study is detailed in the following subsection.

\subsection{Experimental Design} \label{subsec:experimental_design}

We analyzed formative assessment responses from 175 sixth-grade students (ages 11-12) at a public middle school in the southeastern United States. The student population was 67\% White, 14\% Black/African American, 11\% Asian, and 8\% Hispanic/Latino, with three students identifying as other races. The gender distribution was 51\% male and 49\% female. The study protocol was approved by Vanderbilt University's IRB, and all analyzed data were obtained from students who provided informed assent to participate and whose parents provided informed consent. Data were collected over two academic years. Although 175 students participated, some responses were excluded due to non-consent or absences, which resulted in incomplete work. Consequently, 158 students were available for the \textit{Rules Task}, 166 for the \textit{Debugging Task}, and 161 for the \textit{Engineering Task}. The analytic sample, therefore, comprises all consenting, present sixth-grade students from the study site across a two-year period.

During Response Scoring, two of this paper's authors independently scored a randomly selected subset (20\%) of the data using the rubrics described in Section \ref{sec:spice_curriculum}. For each formative assessment, the two humans compared scores and discussed their disagreements before reaching a consensus. Cohen's $\kappa$, the predominant measure in the literature assessing inter-rater reliability between two reviewers, was used as the IRR measure. The Cohen's $\kappa$ values for the \textit{Rules}, \textit{Debugging}, and \textit{Engineering Tasks} were 0.86, 0.74, 0.84, respectively. 

Once consensus was achieved ($\kappa \geq 0.70$), one of this paper's authors scored the remaining instances. Each task's dataset was split into 80/20 training/testing sets before Prompt Development, with training set instances being those considered for inclusion in the prompt as few-shot examples and test set instances used for method evaluation. Instances discussed during IRR were withheld from the test set to prevent data leakage. Instances in the training set not used as few-shot examples in the initial prompt were reserved as a validation set to support the active learning process.

All three task datasets were imbalanced. In the \textit{Rules Task},  the modal total score was a perfect 9/9, occurring in 34 out of 158 cases; $0$ was the second most frequent score observed in 29 instances. Across the six subscores on conditional statements and runoff values, most students earned points; in contrast, for the three absorption-related subscores, students were more likely to receive no credit, indicating a stronger understanding of logic expressions and runoff values than their relationship to absorption. The \textit{Debugging Task}'s distribution was biased towards higher scores, with the mode being a perfect 5/5 for 66 out of 166 students and demonstrating a monotonically decreasing frequency of students attaining lower scores. All five subscores showed students earning points more often than not. The \textit{Engineering Task} was dominated by incorrect responses, with $0$ as the most common score (63\%; 101/161 students); the remaining scores were approximately uniformly distributed from 1 to 4 (inclusive).

To evaluate CoTAL's scoring accuracy on the \textit{Rules}, \textit{Debugging}, and \textit{Engineering Tasks}, we compared the performance of CoTAL to a zero-shot, ``scoring-only'' Baseline (i.e., numerical scores only without labeled instances, CoT reasoning chains, or active learning) to evaluate CoTAL's generalizability, comparing CoTAL-generated scores and explanations to those of the non-prompt-engineered LLM. Performance details for adding each individual component to CoTAL's prompt engineering pipeline are reported in previous work \citep{cohn2024chain}. We used GPT-4\footnote{A temperature of 0 was used to achieve near-deterministic behavior.} to conduct our analysis due to its balance between performance and cost.

To evaluate CoTAL's generalizability for scoring and explaining formative assessment questions across connected domains (RQ1), we employed a mixed-methods design. Quantitatively, we assessed LLM performance on a held-out test set using Cohen's QWK, selected for its prevalence in automated essay scoring \citep{singh2023h}. Unlike Cohen's $\kappa$, Cohen's QWK awards partial credit proportional to the degree of disagreement, making it well-suited to ordinal labels and better aligned with educational rubrics (i.e., it mirrors how human graders value score proximity and partial correctness). Qualitatively, we conducted a constant comparative analysis of GPT-4's scoring explanations and errors to characterize CoTAL's impact on accuracy and feedback utility, and to identify strengths and weaknesses. Implementation details for each formative assessment are provided in the Appendix, and prompts are available in the \href{https://github.com/claytoncohn/CEAI26_Supplementary_Materials}{Supplementary Materials}. We discuss our findings for RQ1 in Section \ref{sec:rq1_findings}.

To investigate student and teacher perceptions of CoTAL's feedback efficacy and its impact on classroom learning (RQ2), we employed qualitative analysis, analyzing teacher interviews and surveying a sample of participating students. First, we conducted two semi-structured interviews with two classroom teachers, each with over five years of SPICE instructional experience and more than 20 years of overall teaching experience. Teachers reviewed LLM responses to previously unseen \textit{Science Concepts and Reasoning} (\ScienceConcepts\ in Figure \ref{fig:spice-assessments}) assessment questions, stated their agreement with the LLM's scores and explanations, and articulated preferences regarding response structure. They also provided recommendations for improving LLM outputs to support students and teachers better.

For students, we conducted a focus group study with 23 consenting students, all present in one classroom during the second year of our two-year study, to evaluate CoTAL's scoring of their \textit{Science Concepts and Reasoning} assessment. The students reviewed the AI-generated scores and feedback and then completed a survey. The survey assessed their agreement with GPT-4's scoring accuracy, its feedback utility, and their confidence in the system's ability to grade future assignments. To answer RQ2, we (1) created memos of key findings from the teachers' interviews and (2) conducted a constant comparative analysis of the students' survey responses. RQ2 results are presented in Section \ref{sec:rq2_findings}.

\section{Analyzing RQ1: CoTAL Generalizability Across Multiple Connected Domains} \label{sec:rq1_findings} 

RQ1 asked, \textit{Can CoTAL improve an LLM's ability to score and explain responses to formative assessment questions across multiple connected domains}? To answer this question quantitatively, we first present performance comparisons between CoTAL and the Baseline for the \textit{Rules}, \textit{Debugging}, and \textit{Engineering} tasks. We then present our qualitative findings, identifying CoTAL's strengths and weaknesses for each task. 

\subsection{Rules Task}\label{subsec:rules_task}

Performance results for the \textit{Rules Task} comparing CoTAL and the Baseline are presented in Table \ref{tab:rules_results}.

\begin{wraptable}{L}{0.4\textwidth}
  \centering
  \small
  \begin{tabular}{|l|ll|}
    \hline
        \textbf{Rule} & \textit{Baseline} & \textit{CoTAL} \\
        \hline
        \textbf{R1} & 0.840 & \textbf{1.000} \\
        \textbf{R2} & 0.936 & \textbf{1.000} \\
        \textbf{R3} & \textbf{0.934} & \textbf{0.934} \\
        \textbf{R4} & 0.467 & \textbf{0.784} \\
        \textbf{R5} & \textbf{0.875} & 0.813 \\
        \textbf{R6} & \textbf{1.000} & \textbf{1.000} \\
        \textbf{R7} & 0.632 & \textbf{0.840} \\
        \textbf{R8} & \textbf{0.934} & \textbf{0.934} \\
        \textbf{R9} & 0.811 & \textbf{0.938} \\
        \hline
        \textbf{Total Score} & 0.930 & \textbf{0.968} \\
        \hline
  \end{tabular}
    \caption{Performance results for CoTAL applied to the \textit{Rules Task} relative to the Baseline in terms of Cohen's QWK. Each ``R'' corresponds to a component subscore for the \textit{Rules Task} and is explained in Table \ref{tab:rules_rubric}. Total Score compares the LLM's prediction of the total score (i.e., the sum of all 9 subscores) to the human label. For each metric, the best-performing implementation is in \textbf{boldface}.}
    \label{tab:rules_results}
\end{wraptable}

% 32 test set instances total
The \textit{Rules Task} Baseline implementation resulted in an average QWK of 0.826 across each subscore.\footnote{Among the metrics we report, the two aggregate measures include: (1) average subscore QWK, calculated as the mean of the QWK values across all rubric subscores (e.g., R1 to R9 for \Rules), i.e., $\frac{1}{n} \sum_{i=1}^{n} \text{QWK}(S_i, \hat{S}_i)$; and (2) Total Score QWK, computed as $\text{QWK}(\hat{T}, \sum_{i=1}^{n} S_i)$, where $\hat{T}$ is the LLM's predicted Total Score, $\sum_{i=1}^{n} S_i$ is the sum of ground truth subscores, and $n$ is the number of subscores defined by the rubric.} Applying CoTAL resulted in an average QWK of 0.916, which represents an average increase of 0.090 (10.9\%) over the Baseline while using CoTAL. By itself, the Baseline implementation provided good results with 7/9 Baseline subscores showing ``strong'' agreement \citep{mchugh2012interrater}. Incorrect predictions were evenly split between false positives and negatives (7 and 6, respectively). The Baseline underscored the Total Score by exactly one point in 7 of the 32 test set instances because it failed to properly add the predicted subscores (e.g., the LLM awarded 8 subscore points but its predicted Total Score was 7). Interestingly, this ``one under'' issue was not observed in previous work \citep{cohn2024chain}. We hypothesize that a large number of subscores (nine for the \textit{Rules Task}) may cause issues during inference due to the exponential nature of combinatorics, resulting in $2^9 = 512$ scoring combinations. Further research is necessary to evaluate this hypothesis empirically (i.e., investigating why the LLM incorrectly summed its own subscore predictions so often). The LLM Baseline incorrectly predicted 17/32 Total Scores overall (12 of which were underscores).

CoTAL improved results, with every individual subscore and Total Score (except R5) beating or tying the Baseline. CoTAL reduced false positives in individual subscores from 7 to 1 while only increasing false negatives by 1. This was explicitly addressed during Active Learning, which favored false positives over false negatives by a 2:1 ratio. R5 was the only subscore that had more than one incorrect prediction with CoTAL, while the Baseline had 4/9 subscores with multiple incorrect predictions. CoTAL resulted in 10 incorrectly predicted Total Scores, compared to 17 for the Baseline implementation, with 8 instances of underscoring and 2 of overscoring. Only 2 of the 10 incorrectly predicted Total Scores were off by two or more points, compared to 5 in the Baseline approach. CoTAL also reduced the Baseline's ``one under'' Total Score versus Baseline addition errors from 7 to 2. 

In addition to accurate scoring, the LLM provided accurate explanations for the scores assigned to the \textit{Rules Task} assessment responses, citing relevant evidence, tying it to specific rubric components, and justifying its scoring decisions like a human scorer. The following example highlights a successful LLM generation for R4 in the \textit{Rules Task}.

\begin{quote}
    \small
    \textbf{Student}: \textit{If rainfall is greater than absorption limit set runoff to absorption limit minus rainfall, and set absorption to absorption limit.} \\

    \textbf{GPT-4}: \textit{The student says `set runoff to absorption limit minus rainfall' inside the `greater than' condition. This is incorrect as the runoff should be rainfall minus absorption limit, not the other way around...}
\end{quote}

In the example, the LLM identifies the pertinent text in the student's response, discerns that the student is incorrect, explains why (using the rubric), and awards the correct score (0). This type of feedback is informative to teachers and students, explaining why a score was awarded instead of merely returning a numerical value. Providing scoring transparency instills more trust in the LLM and automated essay scoring systems if stakeholders understand the rationale behind the LLM's decisions \citep{stewart2023cpscoach}. However, while the LLM's feedback was generally accurate and informative, our experiments show it could be unpredictable and prone to illogical feedback:

\begin{quote}
    \small
    \textbf{Student}: \textit{If rainfall is equal to Absorption limit, set absorption to absorption limit. Set runoff} \\

    \textbf{GPT-4}: \textit{...This is incorrect as the student should set absorption equal to either rainfall or absorption limit in the `equal to' condition per the rubric's guidance...}
\end{quote}

As the example shows, the student correctly identified R4 and set absorption to the absorption limit. The LLM correctly cited the relevant portion of the student's response, acknowledged the applicable rubric directive, yet still failed to award the appropriate point. This is a clear case of \textit{logical inconsistency}, where the LLM output exhibits logical contradictions or is inconsistent with its CoT reasoning chains or cited evidence \citep{huang2023survey}. 

Using CoTAL for the Rules Task showcased the LLM's ability to generate accurate scoring predictions and provide relevant feedback when prompted appropriately. This evaluation illustrates CoTAL's intended purpose: to facilitate scoring and feedback generation based on well-defined formative assessment questions and rubrics. As a result, it allows for a systematic assessment of the LLM's proficiency when given effective prompts.

\subsection{Debugging Task} \label{subsec:debug_task}

The performance comparison between CoTAL and the Baseline for the \textit{Debugging Task} appears in Table \ref{tab:debug_results}. For this task, the LLM needed access to the fictional student's erroneous computational model (see Section \ref{sec:spice_curriculum}), which we distilled into textual form for inclusion in our prompt. Token limitations in GPT-4's context window prohibited Active Learning in the \textit{Debugging Task}, so the results only include the Response Scoring and Prompt Development components of CoTAL Phase II.

\begin{wraptable}{R}{0.55\textwidth}
  \centering
  \small
  \begin{tabular}{|l|ll|}
        \hline
        \textbf{Error} & \textit{Baseline} & \textit{CoTAL} \\
        \hline
        \textbf{D1} & 0.178 & \textbf{0.404} \\
        \textbf{D2} & 0.848 & \textbf{0.926} \\
        \textbf{D3} & 0.374 & \textbf{0.608} \\
        \textbf{D4} & 0.820 & \textbf{0.914} \\
        \textbf{D5} & 0.615 & \textbf{0.678} \\
        \hline
        \textbf{Total Score} & 0.561 & \textbf{0.779} \\
        \hline
  \end{tabular}
    \caption{CoTAL versus the Baseline Performance for the \textit{Debugging Task} using Cohen's QWK. Each ``D'' corresponds to a subscore for the \textit{Debugging Task}, which is explained in Table \ref{tab:debug_rubric}. Total Score compares the LLM's prediction of the total score (i.e., the sum of all 5 subscores) to the human-assigned score. For each metric, the best-performing implementation is in \textbf{boldface}.}
    \label{tab:debug_results}
\end{wraptable}

For the \textit{Debugging Task}, the Baseline implementation achieved an average QWK of $0.567$, whereas CoTAL yielded an average QWK of $0.706$, a 24.5\% improvement. Baseline performance lagged behind the \textit{Rules Task}, with only 2 of 5 subscores attaining QWKs $\geq 0.8$, primarily due to a strong tendency toward false positives (FPs) that outnumbered false negatives (FNs) 19 to 2. Total Score predictions did not exhibit the ``one under'' pattern but aligned less well with human scoring (QWK=$0.561$). The LLM was correct in 23/33 cases, and all 10 errors were due to overscoring (i.e., FPs)---including six instances that were overscored by 2 or more points. Applying CoTAL reduced this imbalance and decreased incorrect predictions across subscores, except for D5. Compared to the Baseline, the LLM made five fewer subscore errors. For Total Score, QWK increased by 0.218 (a 38.9\% gain), and the number of instances with incorrect scores of two or more points declined from 6 to 2, indicating improved alignment with human scorers. 

Under CoTAL, the LLM's feedback explicitly linked cited student text to rubric criteria, prompt guidelines, and the computational model. An example of successful LLM feedback for D1 is shown below.

\begin{quote}
    \small
    \textbf{Student}: \textit{The set absorption limit (inch) of the selected material should be right under set rainfall to 1.} \\

    \textbf{GPT-4}: \textit{...This indicates the student correctly identifies the absorption limit is initially set in the wrong part of the code and that the absorption limit should be set before the first `if' statement...}
\end{quote}

In the example, the LLM isolates the specific portion of the student's response that corresponds to D1. The LLM cites the student, explains why the response earns a point according to the rubric, and provides the correct score (1). This example highlights the LLM's ability to identify relevant portions of the student's response, connect it to the rubric, and assign the correct score. The LLM often referenced other prompt components, such as guidelines and the computational model, citing specific line numbers and code blocks to clarify its scoring decisions. This demonstrates that CoTAL enables the LLM to discern meaning from textual representations of block-based code, even without Active Learning. 

While the LLM generally interpreted the computational model correctly for scoring and feedback, the additional contextual information occasionally impeded information tracking. Consequently, it sometimes misapplied criteria, as illustrated by a case in which it incorrectly awarded a point for D3.

\begin{quote}
    \small
    \textbf{Student}: \textit{In the third rule, it should be rainfall instead of the absorption.} \\

    \textbf{GPT-4}: \textit{The student says, `In the third rule, it should be rainfall instead of the absorption.' The third rule is the `greater than' condition, which is represented in the code by the following block on Line 8: `set [Absorption (inch)] to [Absorption Limit (inch)]'. Therefore, the student correctly identifies that in the `less than' condition, absorption should actually be set to rainfall per the Absorption Set to Absorption Limit rubric criteria...}
\end{quote}

In this case, the LLM correctly identified the ``third rule'' (from the \textit{Rules Task}) as the ``greater than'' condition but erroneously used it to award a point for D3, which pertains to the ``less than'' condition. The LLM linked the ``greater than'' condition to the ``set [Absorption (inch)] to [Absorption Limit (inch)]'' code block on Line 8. However, the ``greater than'' condition did not appear until Line 10. It then cited the D3 rubric and the ``less than'' condition before incorrectly awarding credit. This constitutes \textit{context inconsistency}---outputs unfaithful to the provided context in which the LLM contradicts a fact in the prompt \citep{huang2023survey}---arising from a mix-up of ``rules,'' corresponding ``if statements,'' and line numbers, and indicating difficulty tracking multiple information sources. The example also underscores the interconnectedness of our formative assessments, rubrics, and prompts, which require domain knowledge from the \textit{Rules Task} to assess and provide feedback for \textit{Debugging Task} responses.

Despite some context inconsistency errors, the LLM computed the correct score and provided meaningful feedback for most instances using CoTAL. Additionally, CoTAL demonstrated the LLM's ability to consider textual representations of block-based code. The \textit{Debugging Task} also identified opportunities for refining rubrics and formative assessment questions. For instance, students and the LLM often conflated absorption and absorption limit in the \textit{Rules} and \textit{Debugging Tasks}. This finding can help refine our formative assessment questions, rubrics, and prompts by clarifying this distinction. We hypothesize that instructing students to distinguish between absorption and absorption limit will improve their understanding of science and computing concepts.

\subsection{Engineering Task} \label{subsec:engineering_task}

Unlike the previous \textit{Rules} and \textit{Debugging Tasks}, the \textit{Engineering Task} rubric did not have multiple subscores. Instead, students received a single score in the range 0 to 4 (inclusive) based on the rubric in Table \ref{tab:engineering_rubric}. \textbf{QWK for the Baseline was $0.97$, while QWK using CoTAL was $0.965$}. The \textit{Engineering Task} Baseline resulted in four incorrect scoring predictions out of 32, and the QWK score of $0.97$ was comparable to the agreement between the two humans during IRR (QWK $=0.975$). This supports our previous hypothesis that the human scorer agreement during IRR may indicate the LLM's initial alignment with human consensus \citep{cohn2024chain}. 

CoTAL reduced the number of incorrect predictions from 4 to 2. Although the QWK decreased marginally by $0.005$, this reduction was attributable to a single instance that deviated from the human-assigned score by more than one point. These results highlight two key observations: (1) a decrease in QWK does not necessarily indicate reduced accuracy, as CoTAL achieved higher accuracy with fewer errors; and (2) prompt engineering may yield diminishing returns for tasks where the LLM already performs well out of the box. CoTAL's observed accuracy gain is attributable to Active Learning (Phase II, Step 3 of Figure \ref{fig:method}). Students mentioned specific considerations like ``cost'' and ``accessible squares,'' but the LLM did not connect these to the rubric. After incorporating a validation-set instance into the prompt and introducing corrective CoT reasoning chains, the LLM's accuracy improved.

Consistent with the \textit{Rules} and \textit{Debugging Tasks}, the LLM assigned accurate scores and, using CoTAL, provided constructive feedback by citing evidence from students' responses and aligning it with the rubric. For example, when a student answered ``yes'' in the Engineering Task (incorrect answer; 0 points), the LLM nevertheless offered a constructive critique:

\begin{quote}
    \small
    \textbf{Student}: \textit{Yes. Morgan's second design is better because it's cheaper and has less runoff.} \\

    \textbf{GPT-4}: \textit{...While this indicates the student understands the design constraints (Runoff\_Constraint and Cost\_Constraint, in this case), the student's answer is `Yes'...the student earned a score of 0.}
\end{quote}

This example illustrates that the LLM correctly awarded a score of 0, as the student responded ``yes'' to the assessment question. However, the LLM's feedback also highlighted the student's understanding of key design constraints---cost and runoff---even though no points were earned. This recognition is important: it surfaces evidence of conceptual understanding that might otherwise go unnoticed if only a student's score is considered. This helps teachers acknowledge student understanding and provide them with affirming, formative feedback despite an incorrect response. Without this level of explanation, teachers might overlook the student's grasp of domain concepts.

We also observed instances of undesirable LLM behavior. Consistent with the \textit{Rules} and \textit{Debugging Tasks} (and the \textit{Science Concepts and Reasoning Task} reported in prior work), the LLM was susceptible to misleading responses. In one case, a student wrote, ``\textit{Morgan needs to check how other amounts of rainfall affect her design},'' which the LLM scored as 3 points. However, earning 3 points required identifying the \textit{difference} in rainfall between tests, not the \textit{amounts}. This issue emerged during Response Scoring and was extensively discussed by the research team. Despite emphasizing this distinction in both the rubric and the prompt guidelines, the LLM nevertheless assigned an incorrect score.

A second issue concerned flawed reasoning in the LLM's scoring explanations. For example, a student responded: ``\textit{No. Because she one has better cost and worse absorption, and the other has better absorption and worse cost.}'' The LLM correctly cited the portion of the response referencing engineering constraints, noting it ``\textit{shows an understanding of the trade-offs between the Engineering Constraints}.'' Human scorers consistently assigned this response two points according to the rubric. Nonetheless, the LLM incorrectly asserted, ``\textit{the student does not mention the different rainfall values or the specific Engineering Constraints by name, so the student cannot be awarded 2, 3, or 4 points.}'' This constitutes an \textit{instruction inconsistency} hallucination, wherein the LLM deviates from explicit user instructions \citep{huang2023survey}; notably, the prompt does not require students to identify constraints ``by name'' to receive credit. During Active Learning, we provided CoT reasoning demonstrating that responses referencing trade-offs among absorption, runoff, and cost warrant 2 points. Despite recognizing the student's understanding of these constraints, the LLM failed to assign the correct score.

Overall, the LLM effectively scored student responses and provided clear rationales using CoTAL. As in the \textit{Rules Task}, the \textit{Engineering Task} highlighted the LLM's capacity to assign scores and deliver feedback explicitly aligned with the rubric and student understanding. Whereas this study and prior work primarily employed binary multi-label scoring, the \textit{Engineering Task} further demonstrated the LLM's effectiveness under a multi-class (5-way) scoring scheme. Although CoTAL produced a slight decline in QWK and occasional hallucinations, it enabled the LLM to maintain near-perfect alignment with human scorers and to justify scores accurately using the rubric.

\subsection{Answering RQ1} \label{subsec:answering_rq1}

Overall, CoTAL generalized effectively across tasks and domains, increasing average subscore QWK by 10.9\% and 24.5\% for the \textit{Rules} and \textit{Debugging Tasks}, respectively. In the \textit{Engineering Task}, CoTAL reduced the number of incorrect predictions by two despite a slight decrease in QWK. Using CoTAL, the LLM agreed with researcher evaluations on 94.7\% of the 550 answers across the three formative assessments' test sets, yielding errors on 29 instances in which the model either fabricated information or otherwise produced outputs that diverged from human preferences. These hallucinations were context-dependent, arising from individual misunderstandings rather than domain-specific weaknesses.

Qualitatively, CoTAL generated outputs that accurately justified scores by citing appropriate evidence from student responses and explicitly linking this evidence to the rubrics. While additional mechanisms are needed to reduce hallucinations further, CoTAL improved scoring accuracy and enabled LLMs to provide interpretable scores and explanations across multiple domains and assessments. Each formative assessment required a context-specific prompt (e.g., domain concepts, assessment questions, and rubrics); nevertheless, the same prompt-engineering procedure was effective across science, computing, and engineering without methodological adjustments.

\section{Analyzing RQ2: Teacher and Student Feedback} \label{sec:rq2_findings}

RQ2 asked: \textit{What do teacher and student input reveal about the effectiveness, actionability, and impact of CoTAL's formative feedback}? We answer this question qualitatively by memoing key findings and using constant comparative analysis to analyze teachers' interviews and students' surveys, respectively.

\subsection{Teacher Feedback}

We conducted semi-structured interviews with two classroom teachers, asking them to reflect on CoTAL's scoring of several of their students' \textit{Science Concepts and Reasoning} responses. Both teachers reported that CoTAL achieved high scoring accuracy. They recognized the utility of LLMs in enhancing teaching efficiency and identifying student learning gaps, thereby guiding subsequent educational interventions. One teacher personally undertook the \textit{Science Concepts and Reasoning} assessment and received a score of 6 out of 9 from the LLM. She noted CoTAL's effectiveness in pinpointing and explaining her mistakes and detecting her misunderstanding. The other teacher emphasized CoTAL's (and LLMs', more generally) potential to reduce teacher bias by evaluating students based solely on their answers, not preconceptions about the student:

\begin{quote}
    \small
    \textit{...as a teacher, sometimes you drop the ball because you're like, oh, I know, they meant that, even though they didn't say it...as a teacher, if I'm grading a student's paper that I don't know...I can see all of the things they literally say so much more clearly than if it's a kid I know...that's one great thing about it being done by AI.}
\end{quote}

This teacher also underscored the value of collaboration between humans and artificial intelligence in education, particularly in teaching language models to recognize situations that \textit{necessitate teacher involvement}:

\begin{quote}
    \small
    \textit{...we could train the AI to alert the teacher to that... that's where [the LLM] could help the teacher quickly go, `oh, here's a place to grow this kid's knowledge.'}
\end{quote}

The two teachers proposed refinements to CoTAL, particularly suggesting that the LLM notify the teachers about students who need additional support and provide feedback that recommends subsequent actions to enhance their learning, such as study topics tailored to their knowledge gaps. One teacher outlined three key functions for an enhanced LLM grading system: (1) offering students constructive feedback for reflection, (2) sharing student performance data with teachers, and (3) alerting teachers to notable insights in student submissions. This teacher emphasized the importance of the LLM asking thought-provoking questions to students to evaluate and enhance their conceptual understanding of science topic(s):

\begin{quote}
    \small
    \textit{...that's where the AI could eventually ask an inquiring question...there's your next entry into a discussion.}
\end{quote}

The second teacher expanded on this concept, advocating for the LLM to prompt students to articulate a deep understanding of scientific concepts and interconnections, as opposed to providing surface-level definitions and general overviews of the subject matter: 

\begin{quote}
    \small
    \textit{It's like, okay, you got the big concept. But the little details that make it richer.} \\

    \noindent \textit{For the first student, you would want to know...well, I mean, they said the three different sizes mean three different quantities, but what are those quantities...}
\end{quote}

Both teachers also emphasized the significance of acknowledging student achievements and areas for improvement to ensure that formative feedback encompasses recognition and guidance for further learning. Overall, both teachers were receptive to CoTAL and LLM-guided feedback and were optimistic about LLM use in classrooms going forward. They viewed AI systems as ``tools'' for teachers to improve student feedback and learning, not as replacements for teachers. One teacher stressed the importance of the partnership between teachers and AI, particularly for more routine tasks, as a means of increasing teachers' productivity:

\begin{quote}
    \small
    \textit{...[the AI] doing things that make you more productive as the teacher because that's the kind of stuff that you can do as a teacher, it just is so time intensive, times every kid...so taking some of that legwork out for the teacher...not replacing what the teacher does, just doing the legwork.}
\end{quote}

The teachers focused on two primary ways CoTAL feedback could be effective and impactful for students and teachers: (1) feedback that encourages critical thinking and fosters a deep understanding of concepts, and (2) feedback that alerts teachers to students' misunderstandings and identifies opportunities to expand knowledge. Unlike much of the literature on pedagogical agents, where the \textit{agent} identifies inflection points and performs interventions, both teachers viewed LLMs as the \textit{first step} in the intervention pipeline, identifying inflection points but allowing teachers to decide on feedback. 

However, one teacher noted that it is often impractical for teachers to deliver individual feedback and expressed a desire for CoTAL to provide this feedback in those instances:

\begin{quote}
    \small
    \textit{...and that's where this will be powerful, because giving good feedback is not feasible by the sheer amount you have to give, so it doesn't get given. So finding ways to get more feedback given to kids is where this can come in and be tremendously helpful. Give feedback that...you know, immediate, that you just physically, literally can't do as a teacher on everything.}
\end{quote}

Overall, the teachers view the human-AI partnership as a collaboration where AI can: (1) encourage students; (2) alert teachers to students needing assistance; (3) enable teachers to provide more informed, useful feedback; and (4) offer direct feedback when teachers are unavailable. Unlike students, who addressed several of CoTAL's \textit{shortcomings}, the teachers focused almost exclusively on the \textit{benefits} of CoTAL in classroom settings.

\subsection{Student Feedback}

We conducted a focus group and survey with a class of 23 students to evaluate agreement with GPT-4's scores and explanations generated using CoTAL for the \textit{Science Concepts and Reasoning} formative assessments. We also solicited perceptions of the usefulness of CoTAL's feedback and confidence in AI-based grading for future assignments. Students engaged enthusiastically and provided a range of constructive observations, including both positive and critical perspectives on employing LLMs as automated graders.

Overall, 61\% of students reported that CoTAL's responses were helpful, and 65\% expressed confidence that an LLM would accurately grade future assignments. Many students specifically noted that the LLM's scores and explanations were helpful and accurate, and that the model adequately understood their responses:

\begin{quote}
    \small
    \textit{It is helpful because it explains why I was correct and it helped me to understand my score} \\

    \noindent \textit{I liked that it explained thoroughly [sic] what I did wrong and what I did right.}\\

    \noindent \textit{It understood what I said very well.}
\end{quote}

Several students mentioned they appreciated the LLM's objectivity, calling its responses \textit{honest}, \textit{helpful}, and \textit{not biased} (``\textit{I Ike [sic] how honest it is}''; ``\textit{It was very honest. It was helpful.}''; ``\textit{I liked it was not biased.}''). Even in cases where students disagreed with  CoTAL's scoring decisions (i.e., students felt CoTAL underscored their responses), they still expressed an overall openness to LLM grading by answering ``yes'' to whether or not they trusted AI systems to score future assignments. 

The most frequent comment by students was that CoTAL's description of \textit{how} it awarded (or did not award) points and explaining their errors was helpful, particularly concerning the LLM citing evidence and tying it back to the rubric:

\begin{quote}
    \small
    \textit{It was helpful, because I didn't realize that I wrote rainfall instead of runoff.} % (Student 3)
    \\ \\
    \textit{Yes, it was helpful, and I know where I should improve.} %(Student 4)
    \\ \\
    \textit{...the chat does a relatively good job explaining what I did wrong, I think I could work with the feedback to help me learn the content better.}% (Student 22)
\end{quote}

However, some students were not as appreciative of CoTAL's scoring decisions and explanations, and four students could not list a single thing that CoTAL did well. In general, students' largest complaint was that CoTAL's feedback lacked sufficient detail to improve incorrect answers:

\begin{quote}
    \small
    \textit{I think I would learn better if ChatGPT}\footnote{All formative assessment responses were evaluated using CoTAL with GPT-4, but students often referred to CoTAL as ``ChatGPT'' due to its ubiquity.} \textit{answered the question to show what a better response would be like...I wan't [sic] it to show me what I did wrong and give me a example.} %(Student 13)
\end{quote}

One way to mitigate this issue is to include examples of ideal responses alongside each scoring explanation so the students can compare their responses to the \textit{ground truth} (as established by their teacher) and understand the differences with the rubric. Students also suggested making the LLM's tone \textit{less harsh} and its responses \textit{shorter and less repetitive} (``\textit{Be less harsh.}''; ``\textit{Stop repetition}''; ``\textit{I like how it explains all the things it had for requirements, but its a little long.}'') Like their teachers, students expressed a desire for the LLM to acknowledge their achievements in addition to identifying areas of improvement (``\textit{Showing what the student did right}''; ``\textit{I think it can do better with showing what you did right.}'').

Regarding the effectiveness, actionability, and impact of CoTAL's formative feedback, students emphasized its capacity to \textit{explain} scoring decisions, which helped them pinpoint the misunderstandings underlying incorrect responses. By understanding \textit{why} the LLM assigned particular scores, students were able to map deficiencies in their responses to specific rubric criteria and address corresponding learning gaps. This ``explainability'' appears to foster trust in AI systems; prior work indicates that students are often reluctant to accept AI-generated grades without transparent reasoning, which can reduce their willingness to act on LLM-generated feedback \citep{chan2023students}.

\subsection{Answering RQ2}

Teacher interviews and student surveys yielded several insights into CoTAL's efficacy and classroom utility. Both groups recognized the LLM's scoring accuracy and its capacity to explain why responses were correct or incorrect, thereby identifying learning opportunities and potential interventions. Participants expressed willingness to adopt LLMs in educational contexts, particularly with human oversight, consistent with prior findings that ChatGPT enjoys a supportive attitude in academia \citep{jiang2024widen,strzelecki2024acceptance}. Nevertheless, areas for improvement were noted: 39\% of students did not find the feedback helpful, and 35\% lacked confidence in the LLM's grading capabilities. Some students requested more detailed, diagnostic feedback that pinpointed deficiencies. At the same time, teachers emphasized the need for actionable guidance and ``next steps'' to address learning gaps and deepen understanding of science topics.

These findings suggest concrete refinements to CoTAL. We plan to develop prompts that explicitly highlight correct and incorrect elements of student answers, as recognizing correct components can foster engagement and trust. To support error recognition, we will provide exemplary answers alongside LLM feedback to clarify why full credit was not earned and which concepts were misunderstood. CoTAL will be extended to identify misunderstandings for deeper discussion and to inform teachers so they can support struggling students. We will further strengthen stakeholder involvement through participatory design during CoTAL Phase II (Response Scoring, Prompt Development, and Active Learning).

\section{Discussion} \label{sec:discussion}

Our work provides several implications for K-12 educational methodology and practice. Conventional automated grading approaches, such as supervised fine-tuning, typically demand large annotated datasets and substantial GPU resources, rendering them costly and less adaptable to evolving classroom needs. In contrast, prompt-engineering methods like WRVRT \citep{lee2024applying} do not incorporate input from teachers or students and have not been evaluated comprehensively across multiple domains, limiting their pedagogical alignment and generalizability. CoTAL addresses these limitations by adopting a human-AI approach to aligning LLM behavior with teacher expectations and adapting responses to reflect the preferences of individual stakeholders in a generalizable manner. Crucially, the CoTAL approach achieves this using only a small set of demonstration instances to incorporate human insight, thereby reducing data and compute requirements while preserving cross-domain applicability. These properties position CoTAL as a practical, stakeholder-centered framework for formative assessment in K-12 contexts.

This is particularly valuable for subjective tasks such as feedback generation and formative assessment scoring, where teachers' preferences vary, making a single fine-tuned LLM unlikely to align well with multiple educators. While substantial research has focused on adapting LLMs to personalize \textit{student} interactions, typically via content and feedback tailored to individual learning styles \citep{razafinirina2024pedagogical}, CoTAL addresses a notable methodological gap by personalizing LLMs for educators. By incorporating teacher-specific preferences into scoring and feedback with minimal demonstration instances, CoTAL enables educator-aligned, context-sensitive evaluations that are difficult to achieve through conventional fine-tuning or generic prompting alone.

Teacher and student feedback identified opportunities to strengthen stakeholder alignment and emphasized the necessity for cultivating trust in AI systems as a prerequisite for real-world adoption. Students frequently prefer ChatGPT-assisted (i.e., human-in-the-loop) grading over fully automated approaches \citep{tossell2024student,song2024automated}. Consistent with this preference, \cite{jiang2024widen} reported \textit{general optimism} about AI's potential to enhance teaching among enduring concerns related to ethics, discrimination, and regulatory gaps. \cite{lee2024using} further documented \textit{inappropriate} use cases in which the LLM hallucinated internet material and cautioned that such use could ``\textit{potentially undermine students' practical inquiry skills over time}.'' These findings highlight the need for mechanisms that explicitly prioritize LLM reliability, trust, and curricular alignment---not merely accuracy---a conclusion reinforced by our results.

Hallucination reduction is essential. Huang et al.'s \citeyear{huang2023survey} survey on LLM hallucinations identified several hallucination causes, discussing \textit{reasoning failure} where ``[an LLM] \textit{may struggle to produce accurate results if multiple associations exist between questions},'' even in instances where the LLM possesses the necessary knowledge. In our case, this manifested as context inconsistency errors due to the LLM conflating the absorption and absorption limit concepts in the \textit{Rules} and \textit{Debugging Tasks}. These concepts are inextricably linked and often appear in similar contexts, which challenges differentiation. Just as students struggled to distinguish between these concepts, it is likely that the LLM's training data reflects similar ambiguities, which we will investigate in future work. Hallucination mitigation strategies, such as retrieval-augmented generation (RAG) to ground responses in human-curated data and decoding strategies that prioritize adherence to facts and user instructions, similarly warrant future investigation. 

We hypothesize that as foundation models and LLM methodologies continue to evolve, hallucinations will decline. \cite{dhuliawala2023chain} introduced Chain-of-Verification (CoVe), a reasoning-based approach in which the model generates verification questions from baseline responses to independently validate factual claims and produce corrected, evidence-aligned outputs. Their findings indicate that CoVe reduces hallucinations and improves alignment with task-relevant evidence. Recently, enterprise LLMs have adopted similar strategies, such as the reasoning capabilities integrated into OpenAI's o3 and GPT-5 models.

However, despite the potential for advanced LLMs to reduce hallucinations, our recent work (discussed below) demonstrates that while reasoning-enabled models can help mitigate such errors, CoTAL remains an essential component of our automated scoring and feedback pipeline. Notably, \textit{factual} hallucinations were significantly reduced---that is, LLMs adhered more closely to prompt constraints and appropriately cited evidence from students' assessment responses. Nevertheless, formative assessment grading, evidence elicitation, and feedback generation are inherently subjective tasks. Even with LLM reasoning capabilities, multiple rounds of IRR evaluation, prompt refinement, and few-shot instance selection were necessary during active learning. This underscores the continued need for iterative and systematic prompt engineering for tasks where human agreement is challenging to establish. 

Much of the misalignment between LLM predictions and human consensus in this work stemmed from overscoring (i.e., false positives). Given that LLMs are trained to please humans, it is unsurprising that they tend to award more points than warranted. However, in educational contexts, this behavior risks reinforcing misconceptions that may persist into future learning activities. OpenAI recently rolled back a GPT model update due to sycophantic behavior \citep{openai2025sycophancy}, highlighting the importance of examining this issue in CoTAL. In future work, we will extend our analysis to additional LLMs---including reasoning models and smaller, open-source alternatives--- to better understand where our approach excels or falters across different families of models.

Our work is not without limitations. Although we demonstrated CoTAL's generalizability across science, computing, and engineering within an integrated STEM+C curriculum, additional studies are necessary to evaluate its performance across broader tasks, domains, curricula, and student populations---although we note that our approach has since proven successful when applied to discourse analysis \citep{cohn2024human} and the nursing domain \citep{vatral2024design}. This study assessed CoTAL's scoring performance in a \textit{post hoc} context rather than in real-time classroom use, constraining our ability to infer its impact on student learning and classroom dynamics. Moreover, the human-in-the-loop prompt engineering required by CoTAL can be time-consuming---particularly during active learning---raising questions about scalability and sustainability. Future work must establish CoTAL's capacity to reduce teacher workload and maintain alignment at scale in authentic classroom settings. Recent approaches, such as LLM-as-a-Judge \citep{zheng2023judging} and LLM-as-a-Jury \citep{verga2024replacing} show promise for evaluating LLMs towards this end.

Recently, we conducted a follow-up study in four classrooms, each with approximately 26 students, deploying CoTAL via a \textit{formative assessment agent} using refined formative assessments, rubrics, and prompts reflecting the student and teacher feedback gathered from this study (RQ2). This agent enabled students to engage with their CoTAL-generated formative assessment scores and explanations for the scores, helping them understand their mistakes and identify ways to improve their performance moving forward. While a comprehensive analysis of students' learning gains and agent interactions is forthcoming, we (1) observed substantial time savings with CoTAL compared to our earlier human-scoring efforts---delivering feedback within hours instead of weeks, (2) leveraged this feedback to construct learner models based on evidence derived from the formative assessments \citep{cohn2025theory}, and (3) witnessed a considerable increase in LLM scoring performance across all assessments.

\section{Conclusions}

In this paper, we introduced a novel approach to formative assessment scoring, \textit{Chain-of-Thought Prompting + Active Learning} (CoTAL), which integrates ECD principles, human-in-the-loop prompt engineering, and stakeholder-driven refinement of prompts, assessments, and rubrics. We demonstrated CoTAL's generalizability in scoring and explaining students' responses across various types of assessment questions and rubrics across multiple domains. We presented a common framework for formative assessment development, evaluation, and refinement that significantly enhances both the scoring performance and explainability of LLMs for formative assessments. This collaboration between researchers, educators, students, and AI offers promising avenues for improving teacher interventions, enhancing learning outcomes, and advancing both instructional methods and curriculum design.

% Use the style of numbering in square brackets.
% If nothing is used, default style will be taken.
%\begin{enumerate}[a)]
%\item 
%\item 
%\item 
%\end{enumerate}  

%\begin{itemize}
%\item 
%\item 
%\item 
%\end{itemize}  

%\begin{description}
%\item[]
%\item[] 
%\item[] 
%\end{description}  

% \clearpage %%Remove this from your manuscript

% % Figure
% \begin{figure}%[]
%   \centering
% %    \includegraphics{}
%     \caption{}\label{fig1}
% \end{figure}

% \begin{table}%[]
% \begin{tabular*}{\tblwidth}{@{}LL@{}}
% \toprule
% \midrule
%  & \\
%  & \\
%  & \\
%  & \\
% \bottomrule
% \end{tabular*}
% \end{table}

% Uncomment and use as the case may be
%\begin{theorem} 
%\end{theorem}

% Uncomment and use as the case may be
%\begin{lemma} 
%\end{lemma}



% \bibliographystyle{cas-model2-names}

% \bibliography{references}



% \input{appendix}

\end{document}